# Extreme-SAX: Extreme Points Based Symbolic Representation for Time Series Classification

Muhammad Marwan Muhammad Fuad

Coventry University
Coventry CV1 5FB, UK
ad0263@coventry.ac.uk

**Abstract:** Time series classification is an important problem in data mining with several applications in different domains. Because time series data are usually high dimensional, dimensionality reduction techniques have been proposed as an efficient approach to lower their dimensionality. One of the most popular dimensionality reduction techniques of time series data is the Symbolic Aggregate Approximation (SAX), which is inspired by algorithms from text mining and bioinformatics. SAX is simple and efficient because it uses precomputed distances. The disadvantage of SAX is its inability to accurately represent important points in the time series. In this paper we present Extreme-SAX (E-SAX), which uses only the extreme points of each segment to represent the time series. E-SAX has exactly the same simplicity and efficiency of the original SAX, yet it gives better results in time series classification than the original SAX, as we show in extensive experiments on a variety of time series datasets.

## 1 Introduction

Time Series Classification (TSC) is encountered in several applications ranging from medicine (electrocardiogram, electroencephalogram), finance (stock market, currency exchange rates), to industry (sensor signals) and weather forecast. For this reason, TSC has gained increasing attention over the last decade [2] [6] [7] [8] [18].

Time series data are usually high dimensional, and may contain noise or outliers. Therefore, dimensionality reduction techniques have been proposed as an efficient approach to perform TSC.

Several time series dimensionality reduction techniques have been proposed, of these we mention Piecewise Aggregate Approximation (PAA) [9] [19], Adaptive Piecewise Constant Approximation (APCA) [10], and the Clipping Technique [17].

One of the powerful time series dimensionality reduction techniques is the Symbolic Aggregate Approximation (SAX) [11] [12], which first converts the time series into PAA and then transforms the data into symbols using discretization. Despite its efficiency and simplicity, SAX has a drawback, which is its inability to keep track of important points. Such points are of particular interest in many applications. This is due to the fact that SAX actually applies two dimensionality reductions steps – the PAA and the discretization, without any mechanism to highlight these important points.

In this paper we present a very simple modification of SAX, yet this modification outperforms SAX in TSC task because it focusses on extreme points.

The rest of the paper is organized as follows; In Section 2 we present background on the topic. In Section 3 we present our new method, which we test in Section 4. We draw conclusions and discuss future work in Section 5.

## 2 Background

A univariate time series $T = (t_1, t_2, \ldots, t_n)$ is an ordered collection of $n$ observations measured at, usually equally-spaced, timestamps $t_n$. Time series data are ubiquitous and appear in a wide variety of applications.

Classification is one of the main data mining tasks in which items are assigned predefined classes. There are a number of classification models, the most popular of which is *$k$-nearest-neighbor* ($k$NN). In this model the object is classified based on the $k$ closest objects in its neighborhood. Performance of classification algorithms can be evaluated using different methods. One of the widely used ones is *leave-one-out cross-validation* (LOOCV) - also known by *$N$-fold cross-validation*, or *jack-knifin*g, where the dataset is divided into as many parts as there are instances, each instance effectively forming a test set of one. N classifiers are generated, each from $\mathrm{N}-1$ instances, and each is used to classify a single test instance. The classification error is then the total number of misclassified instances divided by the total number of instances [3].

What makes TSC different from traditional classification tasks is the natural temporal ordering of the attributes [1]. This is why several classification methods have been developed to address TSC in particular.

Applying the Euclidean distance on raw data in TSC has been widely used as it is simple and efficient. But it is weak in terms of accuracy [16]. The use of DTW gives more accurate TSC results, but this comes at the expense of efficiency.

A large amount of research in time series mining has focused on time series representation methods, which lower time series dimensionality, making different time series tasks, such as classification, clustering, query-by-content, anomaly detection, and motif discovery, more efficient.

One of the first, and most simple, time series representation methods is PAA [9] [19], which divides a time series $T$ of $n$-dimensions into $m$ equal-sized segments and maps each segment to a point of a lower $m$-dimensional space, where each point in this space is the mean of values of the data points falling within this segment.

PAA gave rise to another very efficient time series representation method; the Symbolic Aggregate Approximation – SAX [11] [12]. SAX performs the discretization

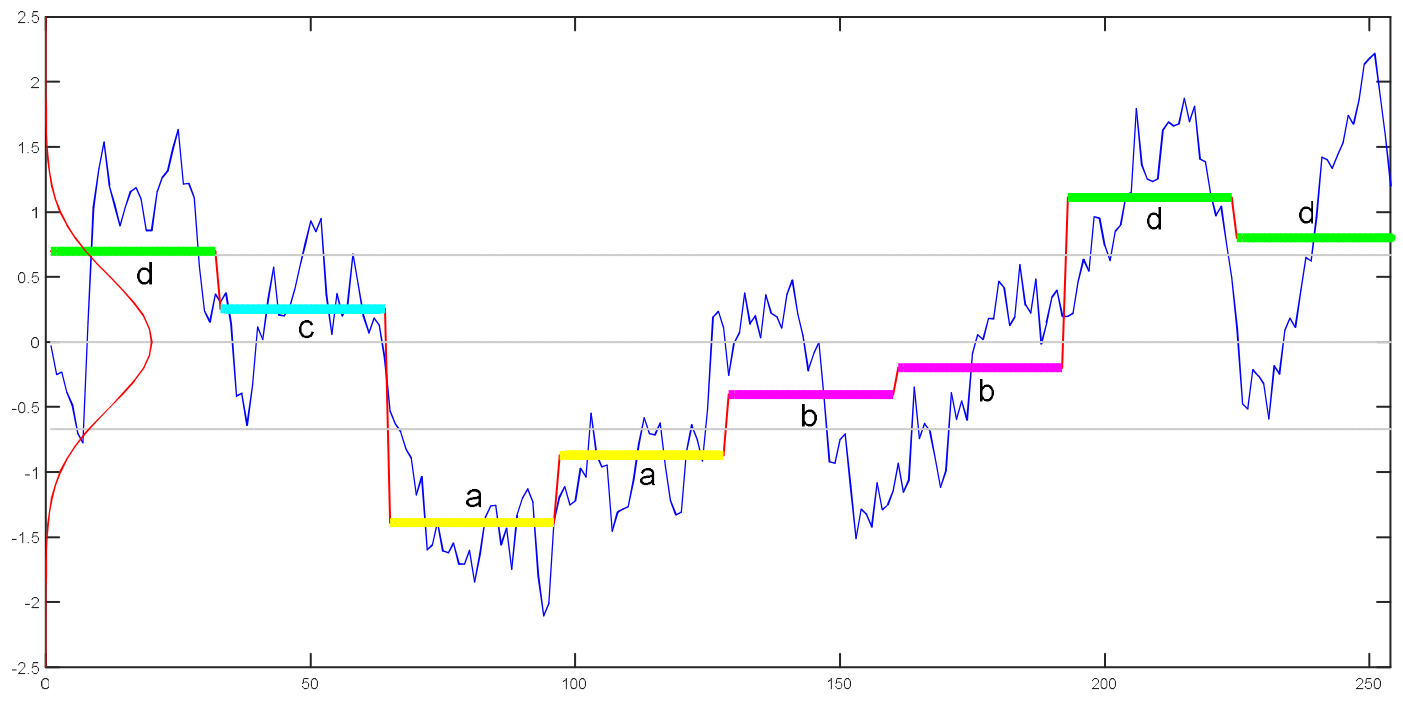

**Fig. 1.** Example of SAX for $alphabetSize = 4, w = 8$. In the first step the time series, whose length is 256, is discretized using PAA, and then each segment is mapped to the corresponding symbol. This results in the final SAX representation for this time series, which is $dcaabbdd$

by dividing a time series $T$ into $w$ equal-sized segments (words). For each segment, the mean value for the points within that segment is computed. Aggregating these $w$ coefficients forms the PAA representation of $T$. Each coefficient is then mapped to a symbol according to a set of breakpoints that divide the distribution space into $alphabetSize$ equiprobable regions, where $alphabetSize$ is the alphabet size specified by the user. Fig. 1 shows an example of SAX for $alphabetSize = 4$

The locations of the breakpoints are determined using a statistical lookup table for each value of $alphabetSize$. These lookup tables are based on the assumption that normalized time series subsequences have a highly Gaussian distribution [12].

It is worth mentioning that some researchers applied optimization, using genetic algorithms and differential evolution, to obtain the locations of the breakpoints [14] [15]. This gave better results than the original lookup tables based on the Gaussian assumption.

Table 1 shows the lookup table for $alphabetSize = 3, \ldots, 10$. Lookup tables for higher values of $alphabetSize$ can be easily obtained.

**Table 1.** The lookup tables for $alphabetSize = 3, \ldots, 10$

| | 3 | 4 | 5 | 6 | 7 | 8 | 9 | 10 |
|---|---|---|---|---|---|---|---|---|
| $\beta_1$ | −0.43 | −0.67 | −0.84 | −0.97 | −1.07 | −1.15 | −1.22 | −1.28 |
| $\beta_2$ | 0.43 | 0 | −0.25 | −0.43 | −0.57 | −0.67 | −0.76 | −0.84 |
| $\beta_3$ | | 0.67 | 0.25 | 0 | −0.18 | −0.32 | −0.43 | −0.52 |
| $\beta_4$ | | | 0.84 | 0.43 | 0.18 | 0 | −0.14 | −0.25 |
| $\beta_5$ | | | | 0.97 | 0.57 | 0.32 | 0.14 | 0 |
| $\beta_6$ | | | | | 1.07 | 0.67 | 0.43 | 0.25 |
| $\beta_7$ | | | | | | 1.15 | 0.76 | 0.52 |
| $\beta_8$ | | | | | | | 1.22 | 0.84 |
| $\beta_9$ | | | | | | | | 1.28 |

## 3 Extreme-SAX (E-SAX)

Despite its popularity and efficiency, SAX has a primary drawback, which is its inability to represent important points accurately. This is due to the loss of information during transformation, first during the PAA stage and second during the discretization stage.

In Fig 2 we see how the two extreme points, represented by red circles in the figure, of the segment were represented by symbol $c$ using SAX with $alphabetSize = 4$, which is clearly not an accurate approximation for these extreme points. This inaccurate representation aggravates in the cases where the accuracy of the representation is mainly based on such extreme points (as in financial time series data, for instance)

In [13] the authors extend SAX to improve its performance in handling financial data. This is done by adding two special new points, that is, max and min points of each segment, therefore each segment is represented by three values; the mean, the min, and max of each segment. Their method uses a modification of the similarity measure that the original SAX uses. The authors say they conducted preliminary experiments on financial time series (the experiments are very briefly discussed in their paper), and they say the results they obtained in a similarity search task is better than those of the original SAX.

The method presented in [13] has a few drawbacks; whereas the original SAX uses one symbol to represent each segment, the method presented in [13] uses three symbols to represent each segment. Using more symbols actually means more information, so it is not surprising that they get better results than the original SAX. Also the similarity measure they apply is slightly more costly than that of the original SAX. In addition, their experiments are quite preliminary, and are applied to financial time series only, so the performance of their method on a large scale of time series datasets of types other than financial is unknown.

In this paper, we present a very simple modification of the original SAX that does not add any additional complexity to it. It still uses one symbol only for each segment, and the similarity measure has the same computational cost as that of the original SAX (which we call *classic-SAX* hereafter). We call our method *Extreme-SAX* (E-SAX).

Let $T$ be a time series in a $n$-dimensional space to be transformed by E-SAX into a $m$-dimensional space, where the size of the word is $w$, i.e. $T$ is segmented into $w$ equal-

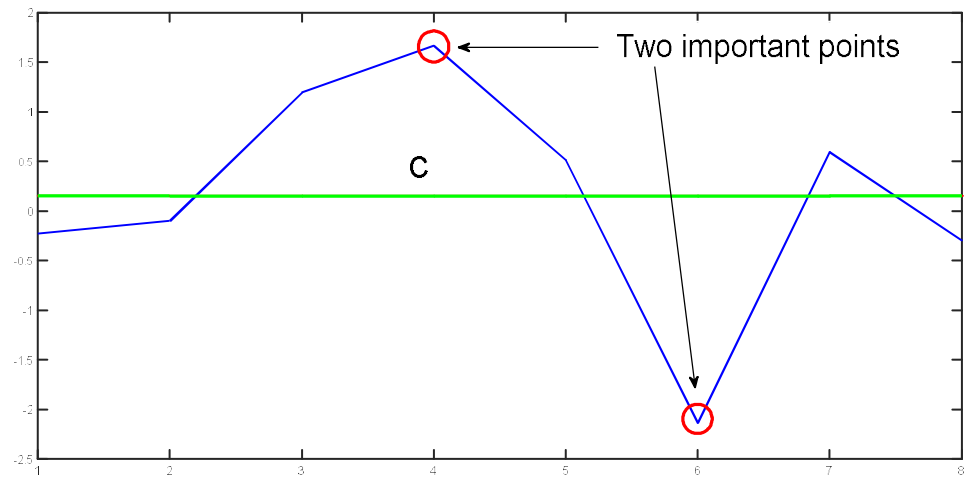

**Fig. 2.** SAX representation for the segment shown in blue, using $alphabetSize = 4$. As we can see, the two extreme points, shown in red circles, are approximated by $c$, which is not an accurate representation.

sized segments. After this first step $T$ will be represented as:

$$T \rightarrow w_1 w_2 w_3 \ldots w_m \quad (1)$$

Let $p_{min}^{i}$ , $p_{max}^{i}$ be the minimum (maximum) of the data points falling within segment $w_i$. We define:

$$p_{mean}^{i} = \frac{p_{min}^{i} + p_{max}^{i}}{2} \quad (2)$$

Using $p_{mean}^{i}$ to represent $w_i$, equation (1) can be written as:

$$T \rightarrow p_{mean}^{1} p_{mean}^{2} p_{mean}^{3} \cdots p_{mean}^{m} \quad (3)$$

In the last step, each coefficient $p_{mean}^{i}$ is mapped to its corresponding symbol using discretization, the same way as in classic-SAX, using $alphabetSize$ symbols.

The distance measure we use in E-SAX is:

$$E - SAX_DIST(\hat{S}, \hat{T}) = \sqrt{\frac{n}{m}} \sqrt{\sum_{i=1}^{m} \left(dist(\hat{s}_i, \hat{t}_i)\right)^2} \quad (4)$$

Where $\hat{S}$ and $\hat{T}$ are the E-SAX representations of the two time series $S$ and $T$, respectively, and where the function $dist(\;\;)$ is implemented by using the appropriate lookup table. This lookup table is the same used in classic-SAX for the corresponding $alphabetSize$.

Unlike classic-SAX, the distance measure defined in equation (4) is not a lower bound of the original distance defined on the $n$-dimensional space. However, this is not important in TSC tasks.

As we can see, E-SAX is very simple. Its symbolic representation has the same length as that of classic-SAX, so it requires the same storage, it does not include any additional preprocessing or post-processing steps, and it uses the same lookup tables as those of classic-SAX.

We can also add that E-SAX clearly emphasizes important points more than classic-SAX. In fact, as we can see, E-SAX representation is completely based on the extreme points of each segment.

## 4 Experiments

We compared the performance of E-SAX to that of classic-SAX in a 1NN TSC task using 45 time series datasets available at the *UCR Time Series Classification Archive* [4]. This archive contains datasets of different sizes and dimensions and it makes up between 90 and 100% of all publicly available, labeled time series datasets in the world,

**Table 2.** Summary of the datasets on which we conducted our experiments

| Dataset | Type | Train | Test | Classes | Length |
|---|---|---|---|---|---|
| synthetic_control | Simulated | 300 | 300 | 6 | 60 |
| Gun_Point | Motion | 50 | 150 | 2 | 150 |
| CBF | Simulated | 30 | 900 | 3 | 128 |
| FaceAll | Image | 560 | 1690 | 14 | 131 |
| OSULeaf | Image | 200 | 242 | 6 | 427 |
| SwedishLeaf | Image | 500 | 625 | 15 | 128 |
| Trace | Sensor | 100 | 100 | 4 | 275 |
| FaceFour | Image | 24 | 88 | 4 | 350 |
| Lighting2 | Sensor | 60 | 61 | 2 | 637 |
| Lighting7 | Sensor | 70 | 73 | 7 | 319 |
| ECG200 | ECG | 100 | 100 | 2 | 96 |
| Adiac | Image | 390 | 391 | 37 | 176 |
| Yoga | Image | 300 | 3000 | 2 | 426 |
| Fish | Image | 175 | 175 | 7 | 463 |
| Plane | Sensor | 105 | 105 | 7 | 144 |
| Car | Sensor | 60 | 60 | 4 | 577 |
| Beef | Spectro | 30 | 30 | 5 | 470 |
| Coffee | Spectro | 28 | 28 | 2 | 286 |
| OliveOil | Spectro | 30 | 30 | 4 | 570 |
| CinCECGTorso | Sensor | 40 | 1380 | 4 | 1639 |
| ChlorineConcentration | Sensor | 467 | 3840 | 3 | 166 |
| DiatomSizeReduction | Image | 16 | 306 | 4 | 345 |
| ECGFiveDays | ECG | 23 | 861 | 2 | 136 |
| FacesUCR | Image | 200 | 2050 | 14 | 131 |
| Haptics | Motion | 155 | 308 | 5 | 1092 |
| InlineSkate | Motion | 100 | 550 | 7 | 1882 |
| ItalyPowerDemand | Sensor | 67 | 1029 | 2 | 24 |
| MedicalImages | Image | 381 | 760 | 10 | 99 |
| MoteStrain | Sensor | 20 | 1252 | 2 | 84 |
| SonyAIBORobotSurface1 | Sensor | 20 | 601 | 2 | 70 |
| SonyAIBORobotSurface2 | Sensor | 27 | 953 | 2 | 65 |
| Symbols | Image | 25 | 995 | 6 | 398 |
| TwoLeadECG | ECG | 23 | 1139 | 2 | 82 |
| InsectWingbeatSound | Sensor | 220 | 1980 | 11 | 256 |
| ArrowHead | Image | 36 | 175 | 3 | 251 |
| BeetleFly | Image | 20 | 20 | 2 | 512 |
| BirdChicken | Image | 20 | 20 | 2 | 512 |
| Herring | Image | 64 | 64 | 2 | 512 |
| ProximalPhalanxTW | Image | 400 | 205 | 6 | 80 |
| ToeSegmentation1 | Motion | 40 | 228 | 2 | 277 |
| ToeSegmentation2 | Motion | 36 | 130 | 2 | 343 |
| DistalPhalanxOutlineAgeGroup | Image | 400 | 139 | 3 | 80 |
| DistalPhalanxOutlineCorrect | Image | 600 | 276 | 2 | 80 |
| DistalPhalanxTW | Image | 400 | 139 | 6 | 80 |
| WordsSynonyms | Image | 267 | 638 | 25 | 270 |

and it represents the interest of the data mining/database community, and not just one group [5].

Each dataset in this archive is divided into two datasets; train and test. The length of the time series on which we conducted our experiments varies between 24 (ItalyPowerDemand) and 1882 (InlineSkate). The size of the train datasets varies between 16 instances (DiatomSizeReduction) and 560 instances (FaceAll). The size of the test datasets varies between 20 instances (BirdChicken), (BeetleFly) and 3840 instances (ChlorineConcentration). The number of classes varies between 2 (Gun-Point), (ECG200), (Coffee), (ECGFiveDays), (ItalyPowerDemand), (MoteStrain), (TwoLeadECG), (BeetleFly), (BirdChicken), and 37 (Adiac). They have a variety of types (simulated, motion, image, sensor, ECG, and spectro). Table 2 shows a summary of the datasets on which we conducted our experiments.

The experimental protocol is as follows; in the train stage each of classic-SAX and E-SAX is applied to each train dataset of the datasets presented in Table 2. The purpose of this stage is to obtain the optimal value of $alphabetSize$, i.e. the value that yields the minimum classification error in TSC for each of the datasets. In the test stage this value of $alphabetSize$ is used in the corresponding test dataset. The final results of TSC on the test datasets for each of classic-SAX and E-SAX are shown in Table 3. The best result (the minimum classification error) for each dataset is shown in boldface printing, yellow-shaded cells.

There are several measures used to evaluate the performance of time series classification methods. In this paper we choose a simple and widely used one, which is to count how many datasets on which the method gave the best performance.

The results show that E-SAX clearly outperforms classic-SAX in TSC as it yielded a lower classification error in 24 datasets, whereas classic-SAX gave better results in 10 datasets only. The two methods gave the same classification error in 11 datasets.

## 5 Conclusion

Classic-SAX is popular time series representation method because of its simplicity and efficiency. It has been widely applied to perform time series tasks such as classification. However, one of its main drawbacks is that it is unable to represent important points accurately.

In this work we presented Extreme-SAX (E-SAX), which uses the extreme points of each segment to discretize the time series. E-SAX has exactly the same advantages of classic-SAX in terms of efficiency and simplicity, but it is better than classic-SAX at representing important points, as it is based completely on the extreme points of each segment to transform the time series into sequences.

We validated E-SAX through TSC experiments on a variety of datasets. Our experiments showed that E-SAX clearly outperforms classic-SAX as it yielded a lower classification error in 24 out of 45 datasets, whereas classic-SAX gave a lower classification error in only 10 datasets. The two methods gave the same classification error in 11 datasets.

For future work, it is worth studying this phenomenon further to know why a representation using less information, based only on the extreme points of segments, gives better results in TSC than a representation that uses more information resulting from all data points of the time series.

**Table 3.** The classification errors of classic-SAX and E-SAX. The best result for each dataset is shown in boldface printing, yellow-shaded cells

| Dataset | Method | |
|---|---|---|
| | classic-SAX | E-SAX |
| synthetic_control | 0.023 | **0.003** |
| Gun_Point | 0.147 | **0.140** |
| CBF | **0.076** | 0.081 |
| FaceAll | 0.305 | **0.275** |
| OSULeaf | **0.475** | 0.484 |
| SwedishLeaf | 0.253 | **0.248** |
| Trace | 0.370 | **0.320** |
| FaceFour | 0.227 | **0.216** |
| Lighting2 | 0.197 | **0.164** |
| Lighting7 | 0.425 | **0.398** |
| ECG200 | 0.120 | 0.120 |
| Adiac | 0.867 | **0.854** |
| Yoga | 0.180 | **0.179** |
| Fish | 0.263 | **0.246** |
| Plane | 0.029 | 0.029 |
| Car | 0.267 | 0.267 |
| Beef | 0.433 | **0.367** |
| Coffee | 0.286 | 0.286 |
| OliveOil | 0.833 | 0.833 |
| CinCECGTorso | 0.073 | 0.073 |
| ChlorineConcentration | 0.582 | **0.508** |
| DiatomSizeReduction | **0.082** | 0.088 |
| ECGFiveDays | **0.150** | 0.235 |
| FacesUCR | 0.242 | **0.206** |
| Haptics | **0.643** | 0.662 |
| InlineSkate | 0.680 | **0.670** |
| ItalyPowerDemand | 0.192 | **0.112** |
| MedicalImages | 0.363 | **0.358** |
| MoteStrain | 0.212 | **0.193** |
| SonyAIBORobotSurface1 | **0.298** | 0.306 |
| SonyAIBORobotSurface2 | **0.144** | 0.146 |
| Symbols | 0.103 | 0.103 |
| TwoLeadECG | 0.310 | **0.278** |
| InsectWingbeatSound | **0.447** | 0.453 |
| ArrowHead | 0.246 | **0.223** |
| BeetleFly | 0.250 | 0.250 |
| BirdChicken | 0.350 | 0.350 |
| Herring | 0.406 | 0.406 |
| ProximalPhalanxTW | 0.370 | **0.362** |
| ToeSegmentation1 | 0.364 | **0.355** |
| ToeSegmentation2 | **0.146** | 0.192 |
| DistalPhalanxOutlineAgeGroup | 0.267 | **0.250** |
| DistalPhalanxOutlineCorrect | **0.340** | 0.398 |
| DistalPhalanxTW | 0.292 | **0.272** |
| WordsSynonyms | 0.371 | 0.371 |
| | 10 | **24** |

## References

1. Bagnall, A., Lines, J., Bostrom, A., Large, J., and Keogh, E.: The great time series classification bake off: a review and experimental evaluation of recent algorithmic advances. Data Min Knowl Disc 31, 606–660 (2017)
2. Baydogan, M., Runger, G., Tuv, E.: A bag-of-features framework to classify time series. IEEE Trans Pattern Anal Mach Intell 25(11):2796–2802 (2013)
3. Bramer, M.: Principles of data mining. Springer (2007)
4. Chen,Y., Keogh, E., Hu, B., Begum, N., Bagnall, A., Mueen, A., and Batista, G.: The UCR time series classification archive. URL www.cs.ucr.edu/~eamonn/time_series_data (2015)
5. Ding, H., Trajcevski, G., Scheuermann, P., Wang, X., Keogh, E.: Querying and mining of time series. Proc of the 34th VLDB (2008)
6. Fawaz, H.I., Forestier, G., Weber, J., Idoumghar, L., Muller, P.A.: Adversarial attacks on deep neural networks for time series classification. In Proceedings of the 2019 International Joint Conference on Neural Networks (IJCNN), Budapest, Hungary, 14–19 July (2019)
7. Hatami, N., Gavet, Y. & Debayle, J.: Bag of recurrence patterns representation for time-series classification. Pattern Anal Applic 22, 877–887 (2019)
8. Karim, F., Majumdar, S., Darabi, H., Chen, S.: LSTM fully convolutional networks for time series classification, IEEE Access 1-7 (2017)
9. Keogh, E., Chakrabarti, K., Pazzani, M. and Mehrotra: Dimensionality reduction for fast similarity search in large time series databases. J. of Know. and Inform. Sys. (2000)
10. Keogh, E., Chakrabarti, K., Pazzani, M., and Mehrotra, S.: Locally adaptive dimensionality reduction for similarity search in large time series databases. SIGMOD pp 151-162 (2001)
11. Lin, J., Keogh, E., Lonardi, S., Chiu, B. Y.: A symbolic representation of time series, with implications for streaming algorithms. DMKD 2003: 2-11(2003)
12. Lin, J., E., Keogh, E., Wei, L., and Lonardi, S.: Experiencing SAX: a novel symbolic representation of time series. Data Min. Knowl. Discov., 15(2), (2007)
13. Lkhagava, B., Suzuki, Y., Kawagoe, K.: Extended SAX: extension of symbolic aggregate approximation for financial time series data representation, Proc. of the Data Engineering Workshop 2006, 2006, 4A0-8 (2006)
14. Muhammad Fuad, M.M.: Differential evolution versus genetic algorithms: towards symbolic aggregate approximation of non-normalized time series. Sixteenth International Database Engineering & Applications Symposium– IDEAS'12 , Prague, Czech Republic,8-10 August, 2012 . Published by BytePress/ACM (2012)
15. Muhammad Fuad, M.M.: Genetic algorithms-based symbolic aggregate approximation. In: Cuzzocrea, A., Dayal, U. (eds.) DaWaK 2012. LNCS, vol. 7448, pp. 105–116. Springer, Heidelberg (2012)
16. Ratanamahatana, C., Keogh, E.: Making time-series classification more accurate using learned constraints. Proc. SIAM Int'l Conf. Data Mining, pp. 11-22 (2004)
17. Ratanamahatana, C., Keogh, E., Bagnall, A. J., and Lonardi, S.: A novel bit level time series representation with implication of similarity search and clustering. In Advances in knowledge discovery and data mining, pages 771–777. Springer (2005)
18. Wang, Z., Yan, W., Oates, T.: Time series classification from scratch with deep neural networks: A strong baseline. In Proc. Int. Joint Conf. Neural Netw. (IJCNN), May 2017, pp. 1578 1585 (2017)
19. Yi, B. K., and Faloutsos, C.: Fast time sequence indexing for arbitrary Lp norms. Proceedings of the 26th International Conference on Very Large Databases, Cairo, Egypt (2000)